\documentclass[10pt, twocolumn, letterpaper]{article}

\usepackage{cvpr}              
\usepackage{graphicx}
\usepackage{amsmath}
\usepackage{amssymb}
\usepackage{epsfig}
\usepackage{multirow}
\usepackage{float}

\usepackage{pifont}
\definecolor{cvprblue}{rgb}{0.21,0.49,0.74}
\usepackage[pagebackref,breaklinks,colorlinks,citecolor=cvprblue]{hyperref}

\def\paperID{10545} 
\def\confName{CVPR}
\def\confYear{2025}

\title{Self-Supervised Large Scale Point Cloud Completion for Archaeological Site Restoration\\[1em] 
    \includegraphics[width=\textwidth]{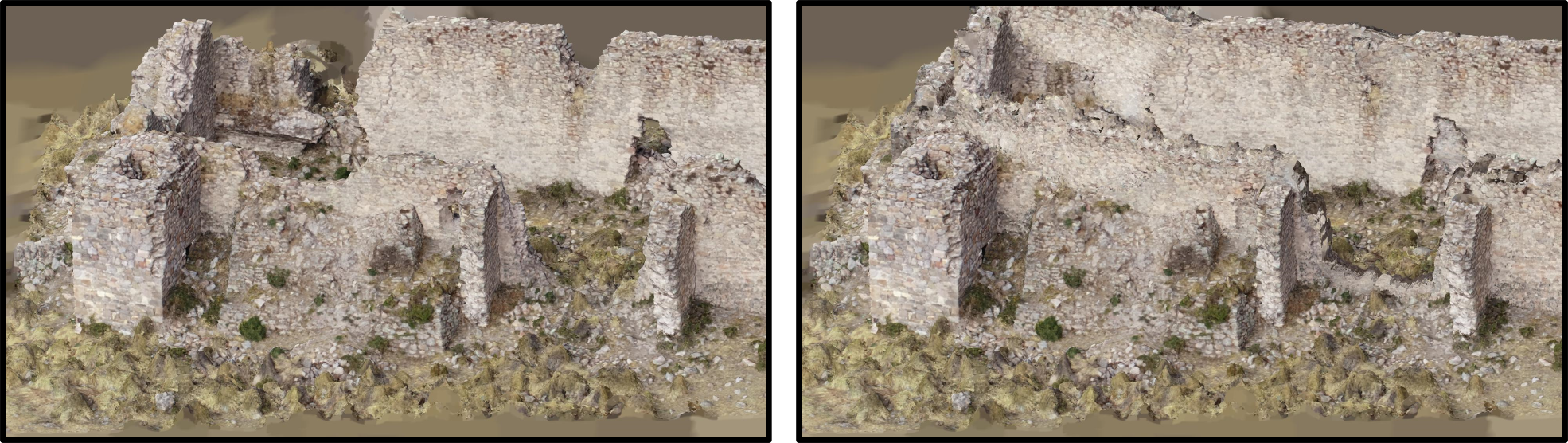}
   \captionof{figure}{Restoration results of applying our method on a large structure consisting of 15 million points from Mawchu Llacta in Peru. The structure is completed and displayed over the original site for visualization.}
   \label{titleFig}
}

\author{Aocheng Li, James R. Zimmer-Dauphinee, Rajesh Kalyanam, Ian Lindsay, \\
Parker VanValkenburgh, Steven Wernke, Daniel Aliaga}

\begin{document}
\maketitle
\begin{abstract}


Point cloud completion helps restore partial incomplete point clouds suffering occlusions. Current self-supervised methods fail to give high fidelity completion for large objects with missing surfaces and unbalanced distribution of available points. In this paper, we present a novel method for restoring large-scale point clouds with limited and unbalanced ground-truth. Using rough boundary annotations for a region of interest, we project the original point clouds into a multiple-center-of-projection (MCOP) image, where fragments are projected to images of 5 channels (RGB, depth, and rotation). Completion of the original point cloud is reduced to inpainting the missing pixels in the MCOP images. Due to lack of complete structures and an unbalanced distribution of existing parts, we develop a self-supervised scheme which learns to infill the MCOP image with points resembling existing "complete" patches. Special losses are applied to further enhance the regularity and consistency of completed MCOP images, which is mapped back to 3D to form final restoration. Extensive experiments demonstrate the superiority of our method in completing 600+ incomplete and unbalanced archaeological structures in Peru.

\end{abstract}    
\section{Introduction}

Point clouds are important for a wide range of vision applications including autonomous driving, real-estate walkthroughs, pre/post-disaster understanding, and digital archaeology and historic preservation (e.g., \cite{hybrid3Drec, pcdForVR, pcdForAutonomousDriving, pcdForMotion}). Effective algorithms for learning and understanding point clouds have been developed and validated on synthetic clean shapes \cite{pointnet++,pointTransformer,pointTransformerV3}. However, for modeling large outdoor real world objects raw point clouds are sparse, noisy, and incomplete. The incompleteness stems from either occlusion (e.g., samples are behind a wall or tree) or missing surfaces (e.g., archaeological remains, post-disaster), which results in different probability distributions of point samples. Hence, point cloud completion for large datasets that include significant missing surfaces/areas, and also lack ground-truth data, is a critical and necessary task. Such 3D reconstructions are vital to understanding the structuring of spaces within archaeological settlements and for modeling social dynamics in the past.

To complete point clouds, prior works \cite{pointpnc, ACLSPC, mal-spc, p2c, chen2020pcl2pcl, cycle4completion} have attempted to fill-in the partial point clouds using supervised, unsupervised, or self-supervised learning methods. Supervised learning rely on the presence of ground-truth and labeled input-output pairs. Unsupervised learning omit labeled input-output pairs but still expect ground truth (e.g., examples of complete models spanning the distribution space). Self-supervised learning does not assume ground truth and generates supervision signals itself (e.g., additional self-masking \cite{pointpnc, p2c}, consistency of multi-view partial inputs \cite{ACLSPC}, or similarity of curvatures \cite{mal-spc}) from fragmented data like KITTI \cite{KITTI-dataset} and MatterPort3D \cite{Matterport3D}. Self-supervision promises to be a good methodology to complete large point clouds datasets with significant missing data and no ground truth (of the missing surfaces).

In this work, we present a self-supervised point-cloud completion approach for large datasets targeted to archaeological settings with a significant amount of missing surfaces due to weathering, erosion, physical stress and human activities \cite{archaeologyDetection}, which would otherwise be extremely hard to restore physically to their original appearance. As opposed to prior methods (e.g.,\cite{pointpnc, ACLSPC, mal-spc, p2c}), we are able to infer point clouds varying from 50k to 2M points, with up to 95\% missing surfaces, and to produce colors/textures. Our approach relaxes the typical uniform-probability assumption of missing points and instead extracts a probability from the partially provided data. This benefits point completion for more general, complex, and large-scale scenarios. 

\begin{figure}[htbp]
  \centering
   \includegraphics[width=\linewidth]{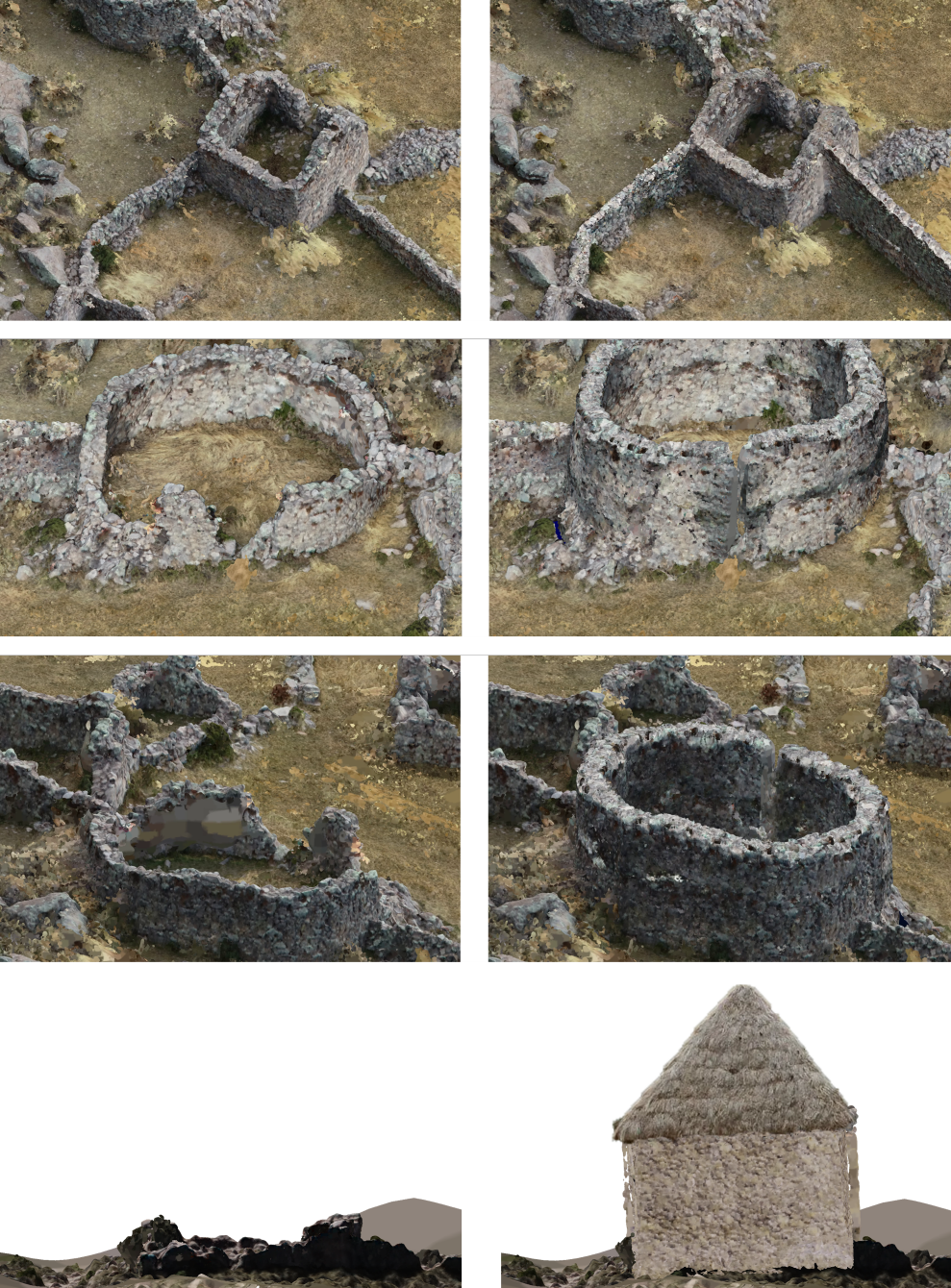}
  
  \caption{Example completion results from Mawchu Llacta and Huamanmarca. The roof in the fourth row is generated with additional synthesis for visualization purposes only.}
  \label{fig:completionResults}
\end{figure}

Our method stores points clouds using multi-center-of-projection (MCOP) images and using patch-based self-supervised inpainting. Instead of processing the point cloud in full 3D, our approach uses an MCOP image to combine points observed from multiple viewpoints along a path. Such images are used to train a network using adversarial loss of sampled patches against complete patches to generate completed patches. The image-based representation scales well to point-cloud completion for millions of points (e.g., see Fig. \ref{titleFig} and Fig. \ref{fig:completionResults}). Further \cite{hybrid3Drec}, we use Poisson Reconstruction \cite{kazhdan2006poisson} on the completed point cloud to produce a 3D visualization. Our experiments show that our method not only out-performs previous self-supervised point cloud completion methods, but also yields finer geometric details than current self-supervised image inpainting methods for similar (archaeological) settings and produces high-quality completion of models with color data, even when far less than half of the point samples are available.

\begin{figure}[htbp]
  \centering
    \begin{subfigure}{1.0\linewidth}
    \includegraphics[width=\textwidth]{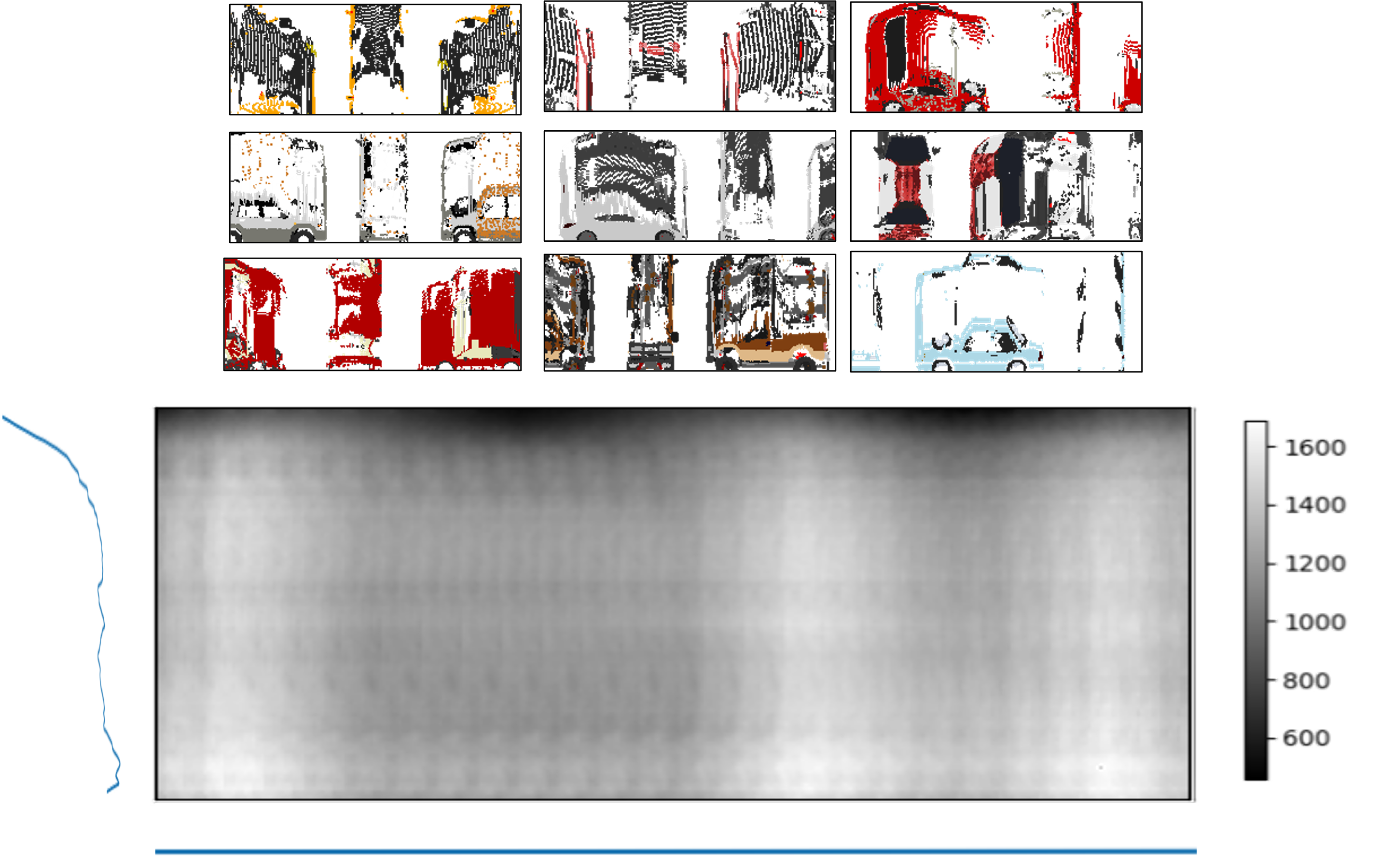}
    \caption{Pixel distribution for ShapeNet cars from partial views.}
    \label{fig:shapenetCarPixelDistribu}
    \end{subfigure}

    \begin{subfigure}{1.0\linewidth}
    \includegraphics[width=\textwidth]{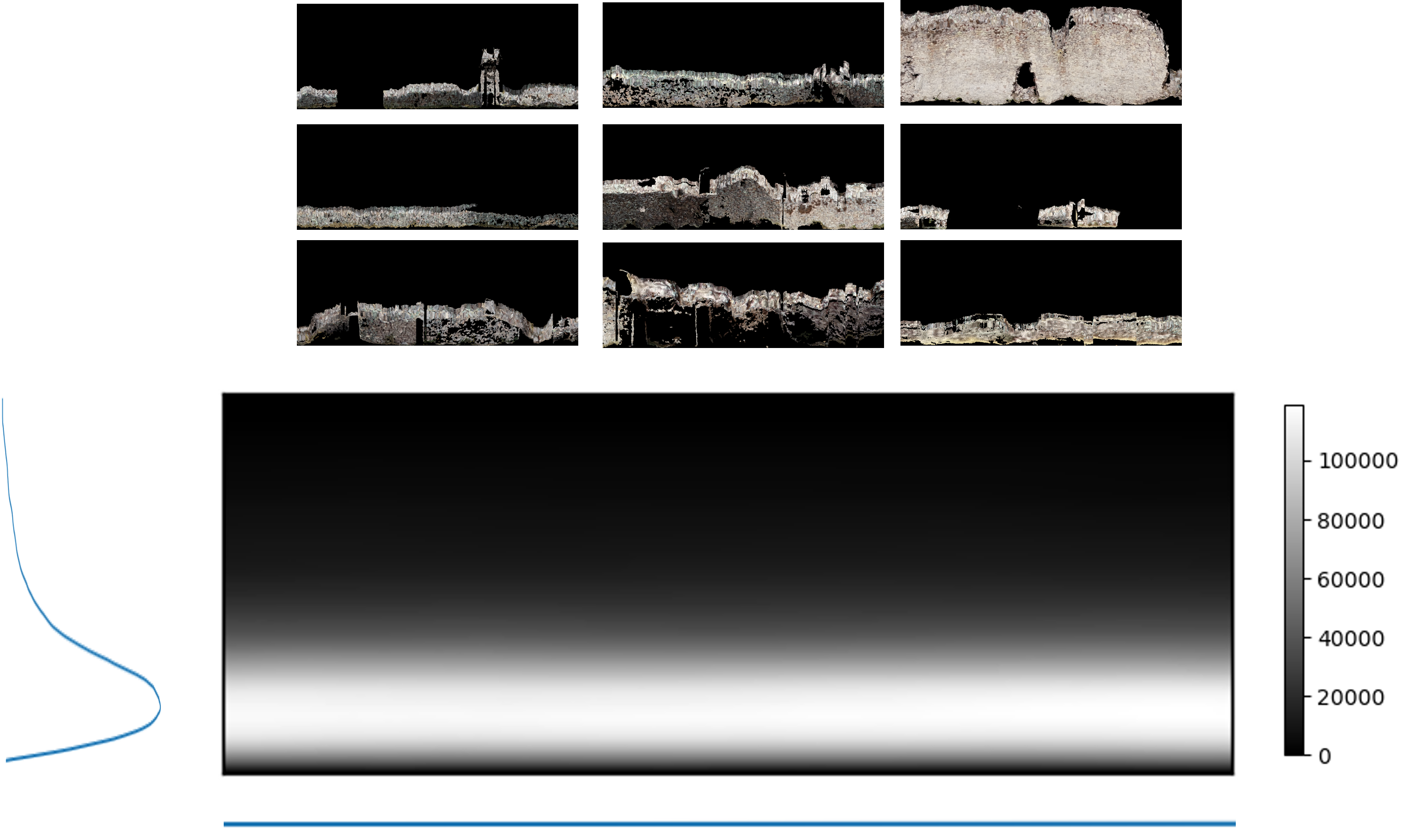}
    \caption{Pixel distribution for our archaeological sites from partial views.}
    \label{fig:missingSurfacePixelDistribu}
    \end{subfigure}
    
  \caption{Comparison of distribution of available points for incomplete point clouds suffering from occlusion/partial view and missing surfaces. The distributions are quite different and that of our outdoor sites is clearly vertically unbalanced.}
  \label{fig:distributionComparison}
\end{figure}

Our main contributions include the following:
\begin{itemize}
    \item[(1)] we provide a novel multiple-center-of-projection (MCOP) representation for point cloud data which unlocks the potential for large scale and high quality point cloud completion of extensive structures,
    \item[(2)] we describe a patch-based point-cloud completion algorithm for a 3D scene encoded into a MCOP image representing millions of points,
    \item[(3)] using (1) and (2) we construct a self-supervised point cloud completion scheme despite spatially unbalanced point distribution and lack of ground truth data (as in the case, at least, in outdoor archaeological settings), and
    \item[(4)] we provide a large-scale 3D point-cloud dataset for improving future methods having to complete acquisitions with an extensive spatial range, millions of RGB points, and high-levels of incompleteness.
\end{itemize}
\label{sec:intro}

\section{Related Works}

\subsection{Supervised/Unsupervised Completion}


Supervised point cloud completion generally requires complete and clean point clouds at training time. Further, it is assumed provided ground-truth data spans the distribution of target objects. Voxel-based methods \cite{dai2017shapecompletionusing3dencoderpredictor, xie2020grnetgriddingresidualnetwork} benefit from 3D convolution but are challenged with processing speed and resolution. Point-based (e.g., \cite{yuan2019pcnpointcompletionnetwork,pointnet++,yang2018foldingnetpointcloudautoencoder,li2023proxyformerproxyalignmentassisted}) can adapt better but neither method works well when ground-truth data are not abundantly provided.
Unsupervised point cloud completion does not rely on paired point clouds. Chen et al. \cite{chen2020pcl2pcl} proposed an early unpaired point cloud completion method using auto-encoders. Other methods (e.g., \cite{cycle4completion,symmetricShapePreservingAutoencoderForUnsupervisedRealScenePointcloudCompletion,zhang2021unsupervised3dshapecompletion,cai2022learningstructuredlatentspace}) have made further improvements. However, this approach still relies on abundant complete point clouds during training and typically fails for uncommon, extensive and/or highly incomplete objects (e.g., \cite{motif, symChineseRec}).

\subsection{Self-supervised Completion}

For some real world scenarios, such as urban environments and archaeological sites, complete samples are costly or unavailable. Self-supervised methods are specifically designed to complete partial point clouds under such settings. Mittal et al. \cite{pointpnc} proposed the first learning-based self-supervised method, PointPnCNet, which randomly masked out small regions during training in order to predict both the removed and the originally missing points. Hong et al. \cite{ACLSPC} developed a closed-loop point cloud completion system named ACL-SPC based on self consistency. 
Cui et al. \cite{p2c} partitioned the partial point clouds into three disjoint groups, which served to improve performance.
Wu et al. \cite{mal-spc} introduced a Pattern Retrieval Network to further find similarities
and subsequently densify the completion. 

These methods pave the way for point cloud completion without ground truth. Nevertheless, they still exhibit several limitations: 1) the size of the point cloud is limited (i.e., 10k or less); 2) the methods generally assume missing data is caused by a limited field of view, which aggregates to a uniform distribution of all the available points. For object completion in outdoor and archaeological settings, the missing data is usually caused by weathering, erosion, physical stress, and activities. This exhibits a heavily unbalanced distribution, which misguides prior methods to overcompletion/undercompletion (see Section 4); 3) the methods usually lack support for color, which is highly important for downstream visualization and analysis applications. Our approach tackles all three of these prior limitations.

\subsection{Computational Archaeology}


Computing is essential to archaeology and to cultural heritage\cite{roleOfComputing}. The Digital Michelangelo project \cite{digitalMichelangelo} paved the way for large-scale archaeological object digitization. Later works addressed completion often by texture/image inpainting \cite{07fragRec}, finding and exploiting symmetries \cite{symChineseRec, kSparseOpt}, and/or a priori object databases \cite{4pointComplet, outRef, hermoza20183dreconstructionincompletearchaeological, sipiran2022dataDriveRestorationOfDigitalArchPott}.

Despite the general success on reassembly of small scale archaeological objects, current methods heavily rely on prior knowledge like different types of symmetries and scanned complete shapes. This does not hold true for large-scale building structures, which also come with almost no available ground truth. However, this is the scenario our work addresses.

\section{Point Cloud Completion}

\begin{figure*}[htbp]
  \centering
   \includegraphics[width=\linewidth]{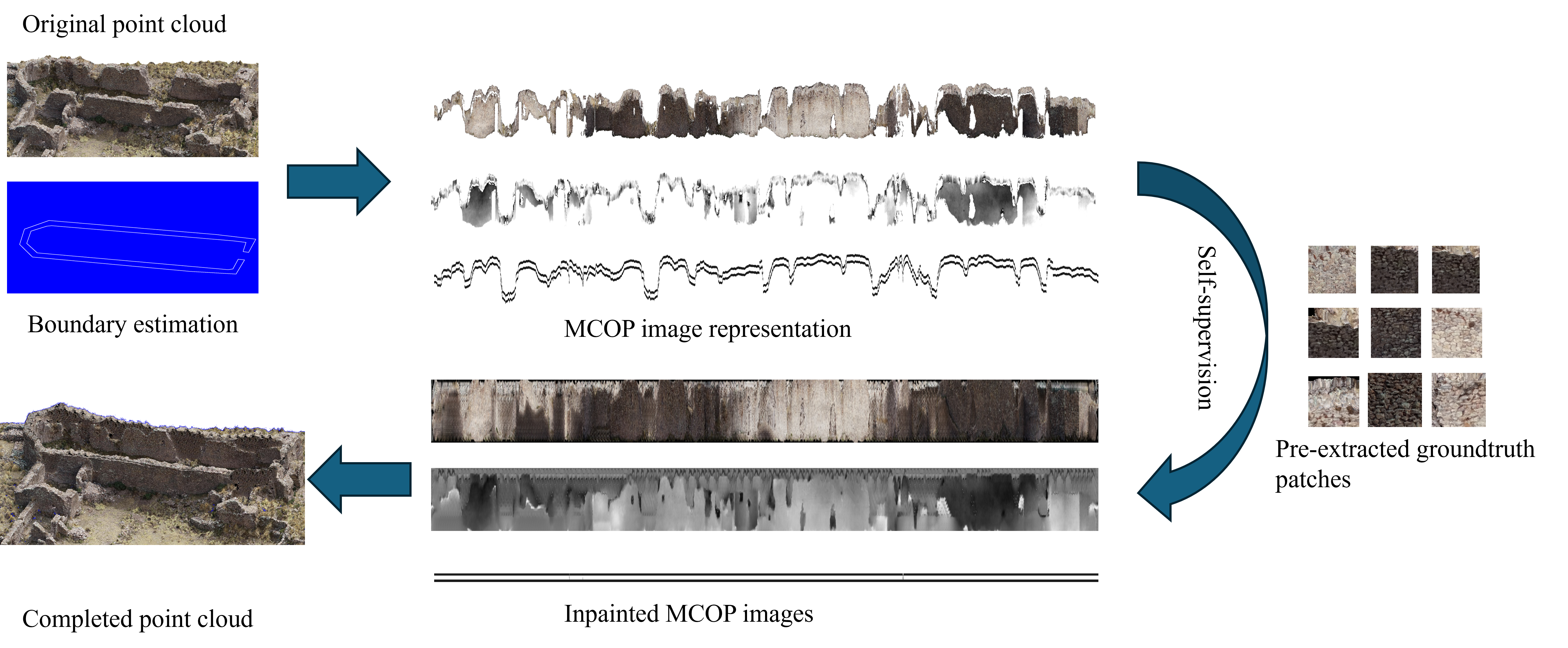}
  
  \caption{General pipeline of our MCOP based point cloud completion method. The input point cloud is first projected onto a 2D image with 5 channels (RGBD + rotation) using our MCOP representation. During training, a random sampling function $f_\theta$ is applied on the MCOP image to extract a group of local windows, which are trained against mostly complete local patches extracted from the dataset of all structures in an adversarial way. For inference, the MCOP image and an outline of the desired shape (e.g., wall height) is passed to our inpainting network adapted from \cite{LaMa}. The final completed point cloud is obtained by remapping the inpainted MCOP image back into 3D.}
  \label{fig:pipeline}
\end{figure*}


For completing a target object, we first project its point cloud $P_{in}$ into 2D using our MCOP representation (Section 3.1). Then, completing the original point cloud is reduced to inpainting the MCOP image $I$ with an inpainter $C$ (Section 3.2), after which the completed shape $P_{out}$ is produced by reprojection $R$ of the infilled MCOP image $P_{out}=R(C(I))$. Finally, Poisson Reconstruction \cite{kazhdan2006poisson} is applied to form a water-tight mesh. An illustration of the whole pipeline is given in Fig. \ref{fig:pipeline}. 


\subsection{Multiple Center of Projection Images}

\begin{figure}[!hb]
  \centering
    \begin{minipage}{0.49\linewidth}
    \begin{subfigure}{1.0\linewidth}
    \includegraphics[width=\textwidth]{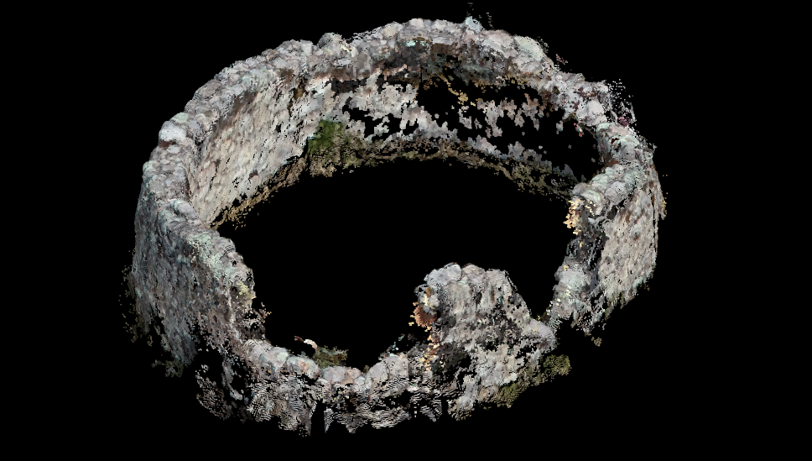}
    \caption{Original structure}
    \end{subfigure}
    \end{minipage}
    \hfill
    \begin{minipage}{0.49\linewidth}
    \begin{subfigure}{1.0\linewidth}
    \includegraphics[width=\textwidth]{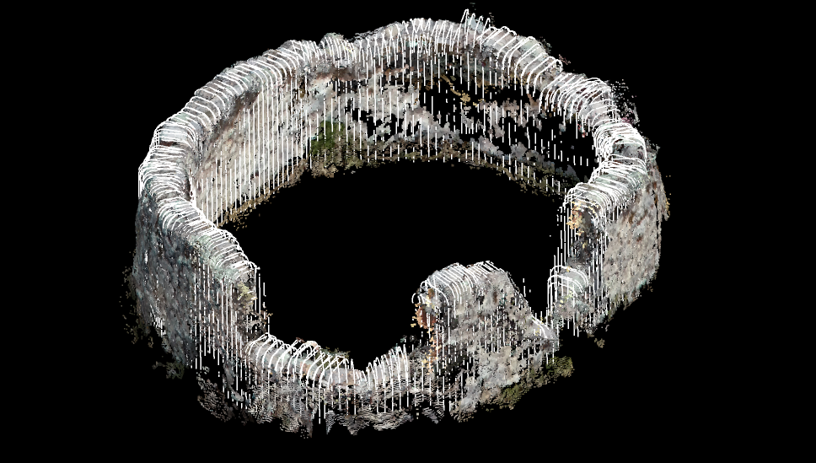}
    \caption{Scan lines}
    \end{subfigure}
    \end{minipage}

    \begin{subfigure}{1.0\linewidth}
    \includegraphics[width=\textwidth]{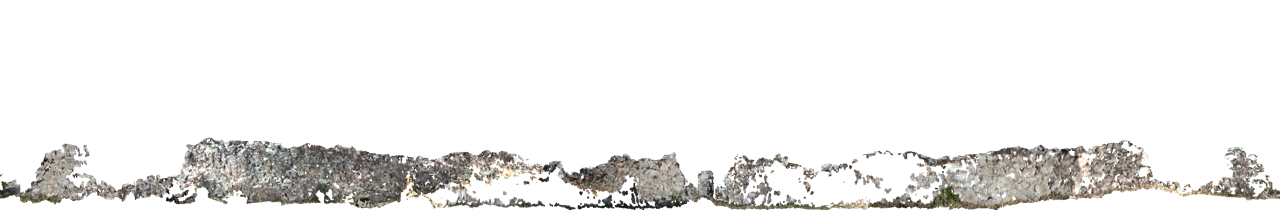}
    \caption{Traditional \cite{MCOP} MCOP Representation}
    \end{subfigure}

    \begin{subfigure}{1.0\linewidth}
    \includegraphics[width=\textwidth]{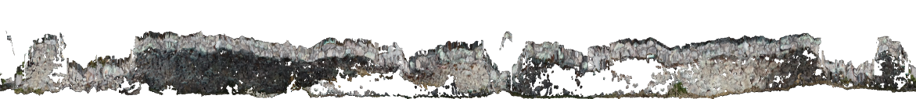}
    \caption{Ours: Color channel}
    \label{mcopStructureSameScanline}
    \end{subfigure}

    \begin{subfigure}{1.0\linewidth}
    \includegraphics[width=\textwidth]{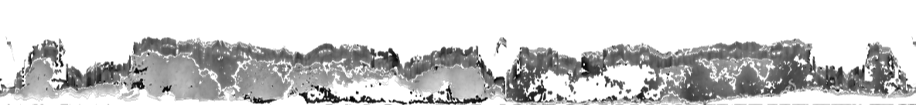}
    \caption{Ours: Depth channel}
    \end{subfigure}

    \begin{subfigure}{1.0\linewidth}
    \includegraphics[width=\textwidth]{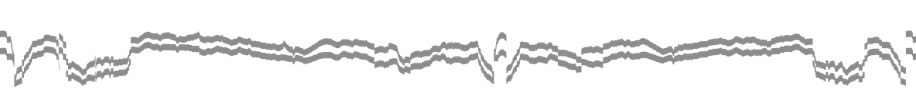}
    \caption{Ours: Rotation channel}
    \end{subfigure}
    
  \caption{MCOP \cite{MCOP} representation. For (a), we sample along a circular path while MCOP camera captures scan lines perpendicular to the path (b). Instead of the traditional MCOP image (c), we capture a 5 channel image (d, e, f).}
  \label{fig:mcopRepresentation}
\end{figure}

To exploit convolutional deep networks ability to process millions of pixels, we construct a single MCOP image $I\in R^{W\times H\times 5}$
to store the raw point cloud data as seen from multiple viewpoints. The object can be represented as an MCOP image in different ways based on the trajectory and pose of the virtual cameras. For a faithful reconstruction of the original point cloud, we argue that by 1) moving the camera along a trajectory around/through the scene and 2) setting the camera to look perpendicular to the trajectory well preserves the scene's fine details. Hence in our work, a virtual camera flies around a target scene/object with its viewing orientation set perpendicular to the flying direction. The pixel-wide columns from each captured image (i.e., "slits") are concatenated to form the MCOP. 



The MCOP image has 5-channels so that each column/slit samples the color and distance to both "side" and "top" of the target object. Unlike the original MCOP representation \cite{MCOP} which captures the side and top into different columns, we include a (viewing) rotation angle so that both parts are continuously placed into a single slit (Fig. \ref{mcopStructureSameScanline}).


The rotation channel also enables shape manipulation. As shown in supplemental, during generation the user can specify different rotations to essentially change the wall height or overall shape, enabling restoring and completing the archaeological structures based on optional simple user guidance.

Formally, we represent the total process as 
\begin{equation}
    I, M = MCOP(P_{in}, \mathbf{C})
\end{equation}
where $\mathbf{C}=\{C_j|j\in[1, H]\}$ are the cameras used for each column of pixels, $I\in R^{W\times H\times 5}$ is the MCOP image, and $M \in R^{W\times H}$ is a binary mask defined by
\begin{equation}
    M_{ij}=\left\{
    \begin{array}{cc}
       0  & \mathrm{if\qquad}I_{ij4}=\infty; \\
       1  & \mathrm{otherwise.} 
    \end{array}\right.
\end{equation}
which is zero for pixels which do not intersect any point by camera $C_j$ at position $(i, j)$.

\subsection{Self-supervised MCOP Completion}
\begin{figure}[htbp]
    \centering
    \includegraphics[width=\linewidth]{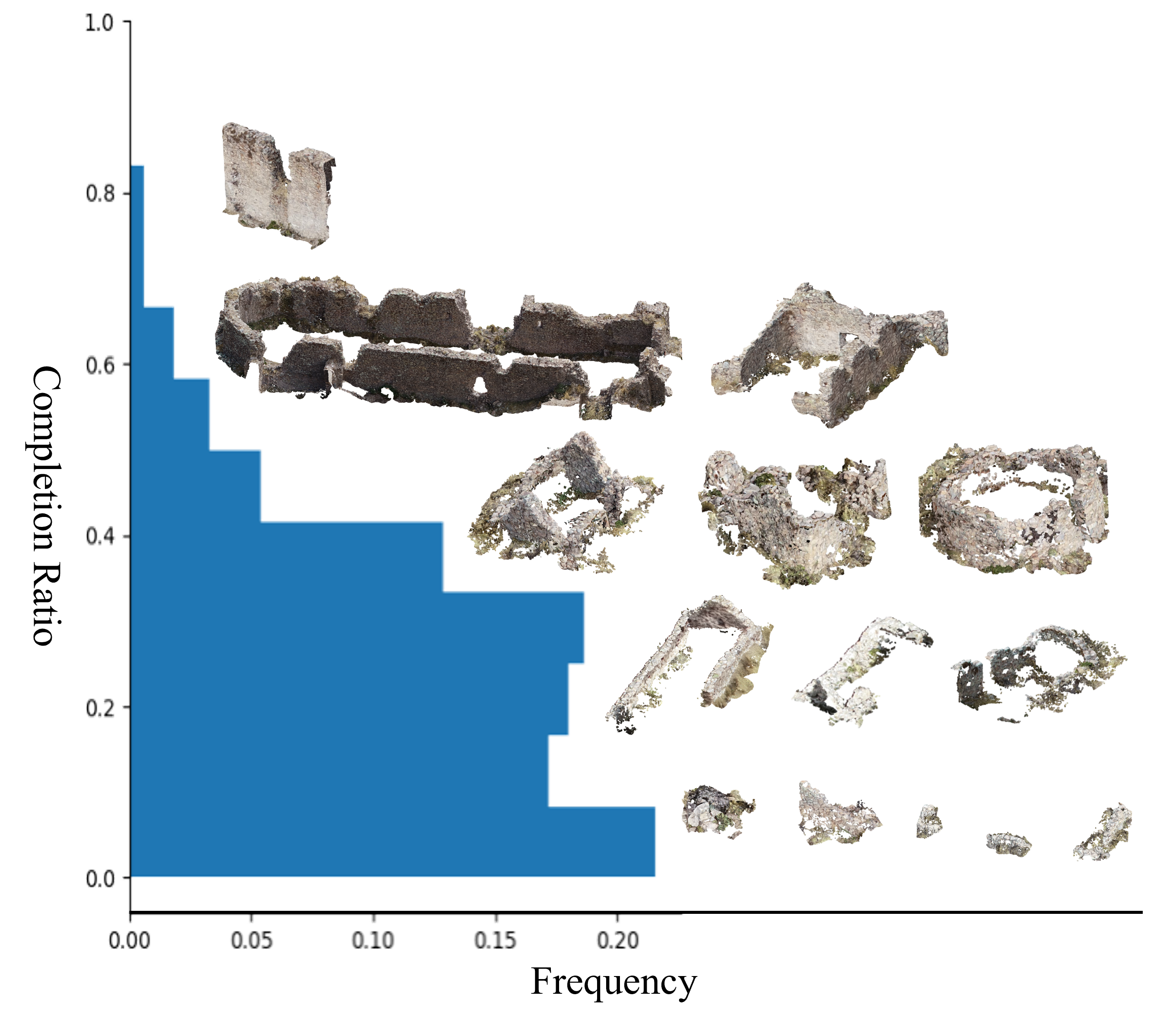}
    \caption{Histogram of random point cloud samples from the Mawchu Llacta dataset, with high completion ratios at the top to low-completion ratios at the bottom.}
    \label{fig:inputDistribution}
\end{figure}

Using MCOP image $I$, the completion for an object fragment is achieved through infilling the missing pixels. We tackle the scenario of high-levels of incompleteness (e.g., over $90\%$ of the walls are missing over $90\%$ of pixels) and lack of ground truth. This results in the sampled point distribution exhibiting a non-uniform distribution (e.g., see Fig. \ref{fig:missingSurfacePixelDistribu} and Fig. \ref{fig:inputDistribution} -- most points are missing in the top). These problematic features still manifest in the MCOP images, for which we resort to a self-supervision mechanism which finds "signals" in subsets of the MCOP images to use in self-supervision during training.

Image generation and completion in self- and un-supervised approaches have been addressed in the literature \cite{ambientGAN, misgan, unsupervisedImageInpainting, SSII}. In these approaches, given partial observation of MCOP images $\mathbf{D}=\{I_1, I_2, \ldots, I_n\}$ and corresponding mask dataset $\mathbf{M}=\{M_1, M_2,\ldots, M_n\}$ with $I_i, M_i=MCOP(P_i, \mathbf{C}), i=1,\ldots,n$, an observation function $f_{\mathbf{M}}:R^{W\times H\times 5}\rightarrow R^{W\times H\times 5}$
\begin{equation}
    f_\mathbf{M}(x) = M_k\otimes x
\end{equation}
with $M_k$ drawn at random from $M$ would be applied on the completer output $O$ to generate a masked sample of the completion with the masked pixels in $x$ filled with a given constant value (i.e., 0 in our case). In \cite{unsupervisedImageInpainting}, the completer $C$ is then trained adversarially to complete images from $\mathbf{D}$ to $\mathbf{D^\prime}=\{I^\prime_1, I^\prime_2,\ldots, I^\prime_n)\}$ through 
\begin{equation}
    I^\prime_i = C(I_i)
\end{equation}
by making the masked observation of $\mathbf{D^\prime_{\mathbf{M}}}=\{f_\mathbf{M}(I^\prime_1), f_\mathbf{M}(I^\prime_2), \ldots, f_\mathbf{M}(I^\prime_n)\}$ to be indistinguishable from $\mathbf{D}$. 

However, directly applying the above would fail under our targeted scenario where the distribution of available pixels in the image is heavily unbalanced (Fig.~\ref{fig:missingSurfacePixelDistribu}). Because few fragments have the upper part of the wall, the completer $C$ will learn to escape from adversarial penalty by simply filling the upper part with a default value, as shown in Fig. \ref{fig:patch level:uair}. It will not be heavily penalized under the masked observation of $D^\prime_M$. 

Instead, to provide effective supervision on both pixels from bottom and top, we apply the following patch-based observation function $f_{w}$ 
\begin{equation}
    f_{w}(I, x, y) = I[x:x+w, y:y+w]
\end{equation}
where $x$ and $y$ are the coordinates for the upper left corner of the window and $w$ is the fixed size of the patches. Despite the sparseness and unbalanced data in the MCOP images, there are smaller windowed samples of uniform completeness. By carefully selecting the observation window size $w$, we extract a new observation dataset $D_w$ composed of all such patches under observation $f_w$. One choice would be $f_{64}$ with windows of no less than $95\%$ completeness, whose average distribution will be (nearly) uniform, and enable the self-supervised strategy in  \cite{SSII, unsupervisedImageInpainting}.

We empirically search for the optimal patch size $w$ and construct the patch ground truth dataset $\mathbf{D}_w$ based on the data distribution. For each MCOP image $I$, we use a high resolution image inpainting network derived from LaMa \cite{LaMa} to obtain the output $O$ through $O=C(I)$. To encourage the completion to be genuine, a random patch $o$ observed by
\begin{equation}
    o = f_w(O, x_i, y_i)
\end{equation}
with $x_i$ and $y_i$ sampled from the image coordinates uniformly should be in the distribution modeled by $\mathbf{D}_w$. The observation patches are fed into the discriminator $D$ adapted from \cite{projectedGAN} to train adversarially against random patches drawn from $D_w$ to minimize the adversarial loss
\begin{equation}
    L_{adv} = \mathbb{E}_{I^\prime\sim D^\prime}[(1-\log(D(f_w(C(I^\prime)))]
\end{equation}
while discriminator $D$ is optimized by
\begin{equation}
    L_{D} = \mathbb{E}_{I\sim D_w}[D(I)] + \mathbb{E}_{I^\prime\sim D^\prime}[(1-\log(D(f_w(C(I^\prime)))]
\end{equation}
\textbf{Regularity Terms.} To further improve the quality of the generated MCOP images, additional loss terms are applied. To drive completion to reproduce the input MCOP image properly, consistency loss is applied on the known pixels as
\begin{equation}
    L_{cons} = ||(O-I)\otimes M||_2 = ||(C(I)-I)\otimes M||_2
\end{equation}
which encourages the completion to reproduce existing MCOP regions. To correctly reproject $O$ to a closed 3D object, each column of pixels in the rotation channel of $C$ should sum up to the angle camera rotated, giving
\begin{equation}
    L_{reg} = (\sum_{j=1}^{h}O_{ij} - \pi)^2
\end{equation}
\textbf{Appearance Consistency Terms.} Due to the relative limited field of view of observation patches compared to the original image size, additional effort has to be taken to ensure the consistency of patterns and textures at a larger receptive field. Inspired by texture synthesis works \cite{gatys2015texturesynthesisusingconvolutional}, we use the pretrained VGG19 network to extract deep features on adjacent patch pairs randomly extracted from completion $O$ and minimize their correlations of features in order to yield a similar appearance. The term is defined as
\begin{equation}
    L_{sim} = \sum_{ij} ||G(f_w(O, x_1, y_1)) - G(f_w(O, x_2, y_2))||_2
\end{equation}
where $G$ computes the Gram matrix of deep features. The optimization goal is defined as
\begin{equation}
    L = L_{adv}+\lambda_{cons}L_{cons}+\lambda_{reg}L_{reg}+\lambda_{sim}L_{sim}
\end{equation}
where $\lambda_{cons}, \lambda_{reg}$ and $\lambda_{sim}$ are weights controlling the importance of different terms.

\subsection{Implementation Details}

Based on the anticipated maximum height of structures, we generally divide the objects to complete into resolutions of $R^{256\times W\times 5}$ and $R^{384\times W\times 5}$, which maximizes the pixel utility in the MCOP image for completion of $4$ and $6$ meter maximum height structures. Two different types of rotation channels are also preconstructed by making the rotation at the expected height and fed jointly with the original RGBD channels from the MCOP image. Noticing the influence of $f_w$ on different resolutions, we apply $\mathbf{D}_{64}$ and $\mathbf{D}_{96}$ separately on the above two distributions. To mitigate for efficient training with abundant self-supervision signals, we precomputed the necessary patch statistics and store the desired patches as an efficient way of accessing $\mathbf{D}_w$.

To train the high resolution image inpainter $C$, we use the proposed self-supervised scheme on MCOP images with a batch size of $4$ for $1$ million iterations. We use the Adam optimizer with an initial learning rate of $0.0003$ and decreasing by $2\%$ every 600 iterations. The model is trained on 4 NVIDIA V100 GPUs with NV-LINK for 3 days.

With the model properly trained, we feed the MCOP images of the target objects into $C$ to get the completed MCOP representation $O$, which is back projected into a 3D point cloud. To mitigate for minor issues and facilitate downstream processing/visualization, we apply Poisson Reconstruction on the output to obtain a watertight mesh with colors as texture.

\section{Experiments}

\begin{figure*}[htb]
\centering
    \begin{minipage}{0.13\linewidth}
    \begin{subfigure}{0.9\linewidth}
    \includegraphics[width=\textwidth]{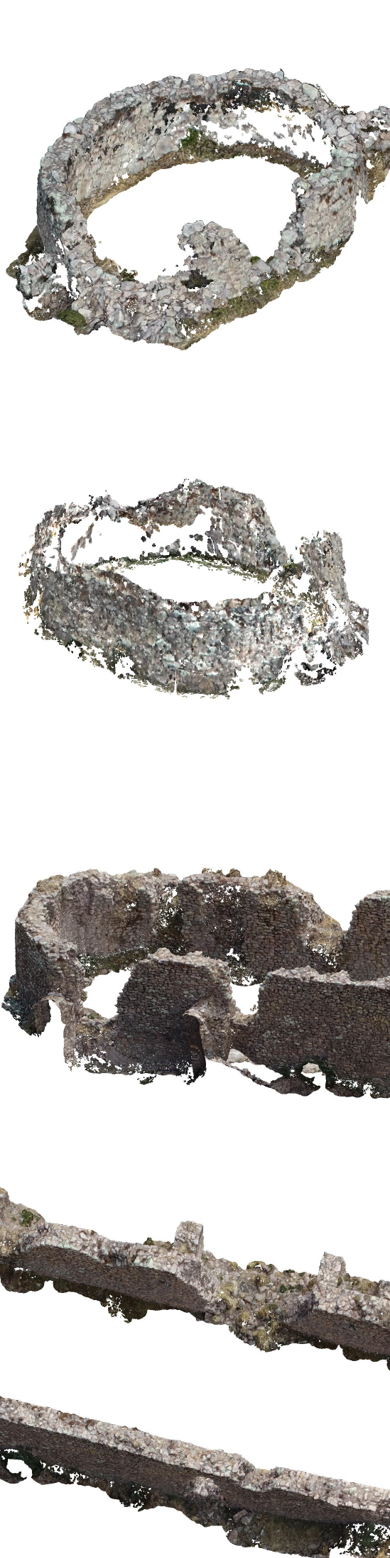}
    \caption{Inputs}
    \end{subfigure}
    \end{minipage}
    \begin{minipage}{0.13\linewidth}
    \begin{subfigure}{0.9\linewidth}
    \includegraphics[width=\textwidth]{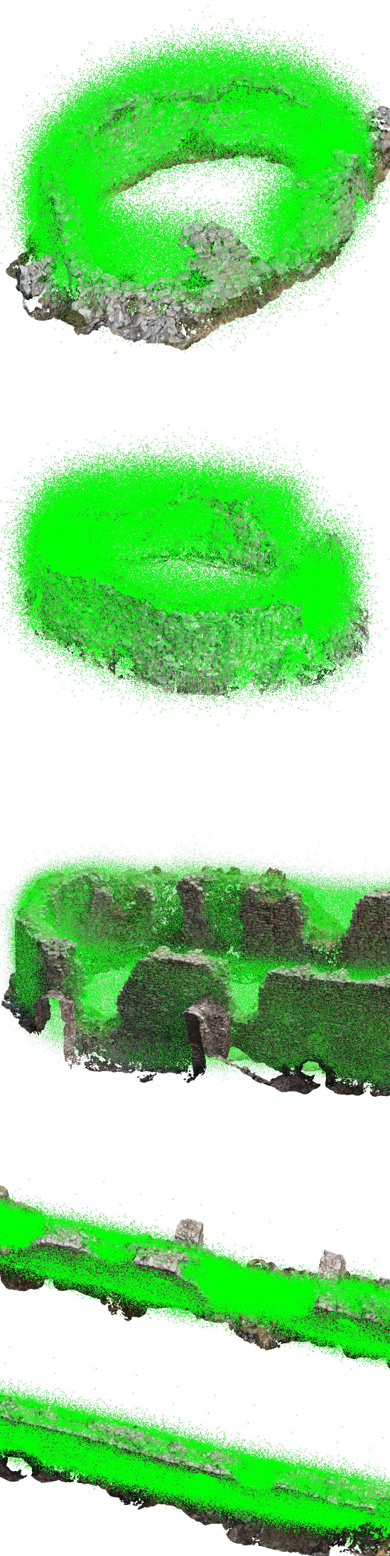}
    \caption{P2C}
    \end{subfigure}
    \end{minipage}
    \begin{minipage}{0.13\linewidth}
    \begin{subfigure}{0.9\linewidth}
    \includegraphics[width=\textwidth]{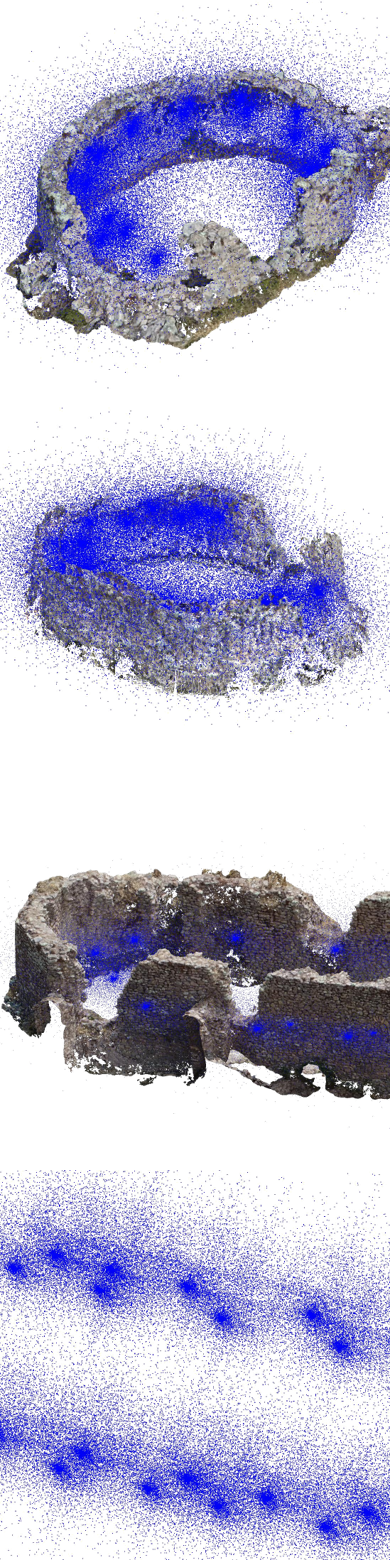}
    \caption{ACL-SPC}
    \end{subfigure}
    \end{minipage}
    \begin{minipage}{0.13\linewidth}
    \begin{subfigure}{0.9\linewidth}
    \includegraphics[width=\textwidth]{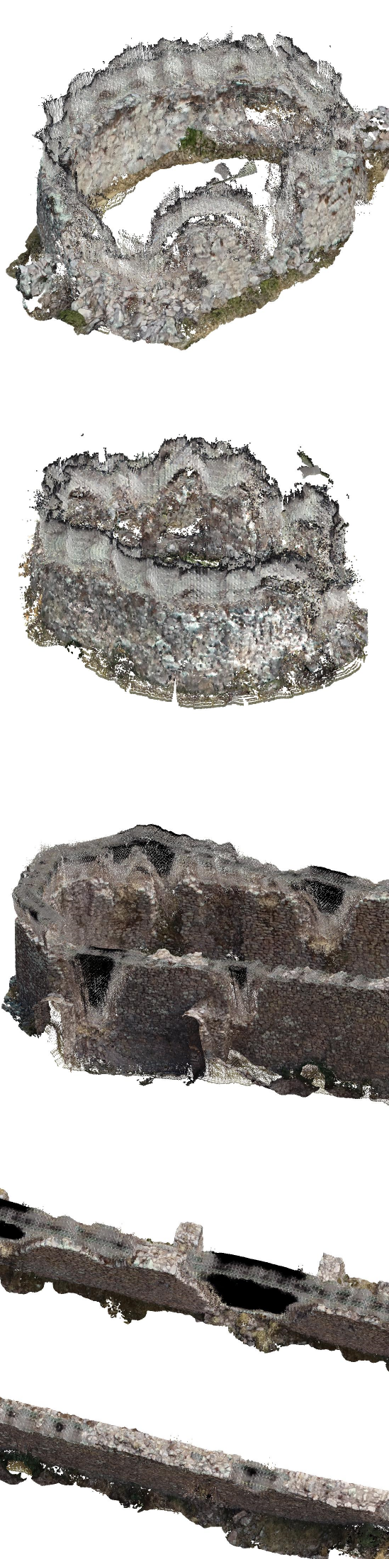}
    \caption{UAIR}
    \end{subfigure}
    \end{minipage}
    \begin{minipage}{0.13\linewidth}
    \begin{subfigure}{0.9\linewidth}
    \includegraphics[width=\textwidth]{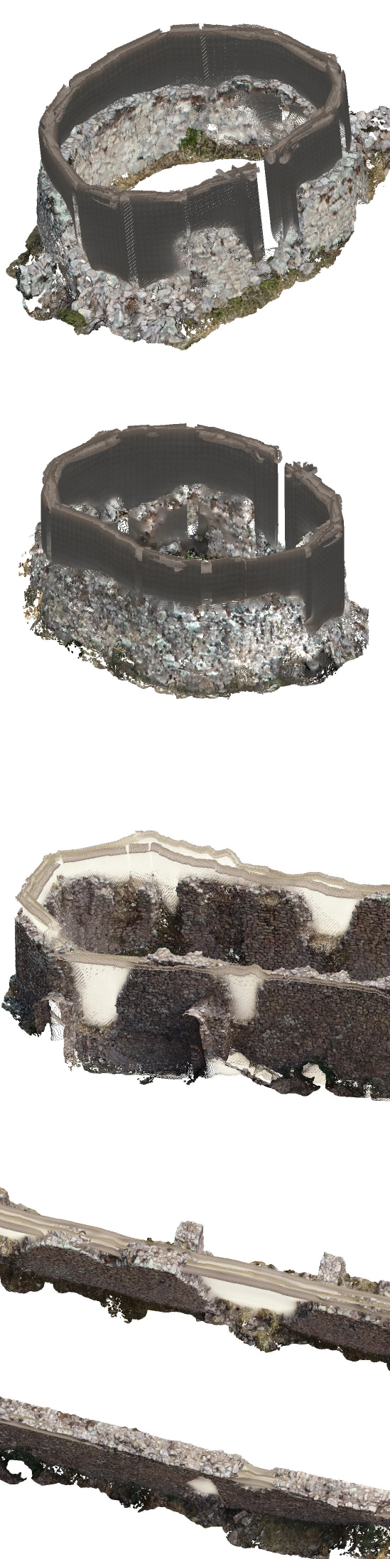}
    \caption{SSII}
    \end{subfigure}
    \end{minipage}
    \begin{minipage}{0.13\linewidth}
    \begin{subfigure}{0.9\linewidth}
    \includegraphics[width=\textwidth]{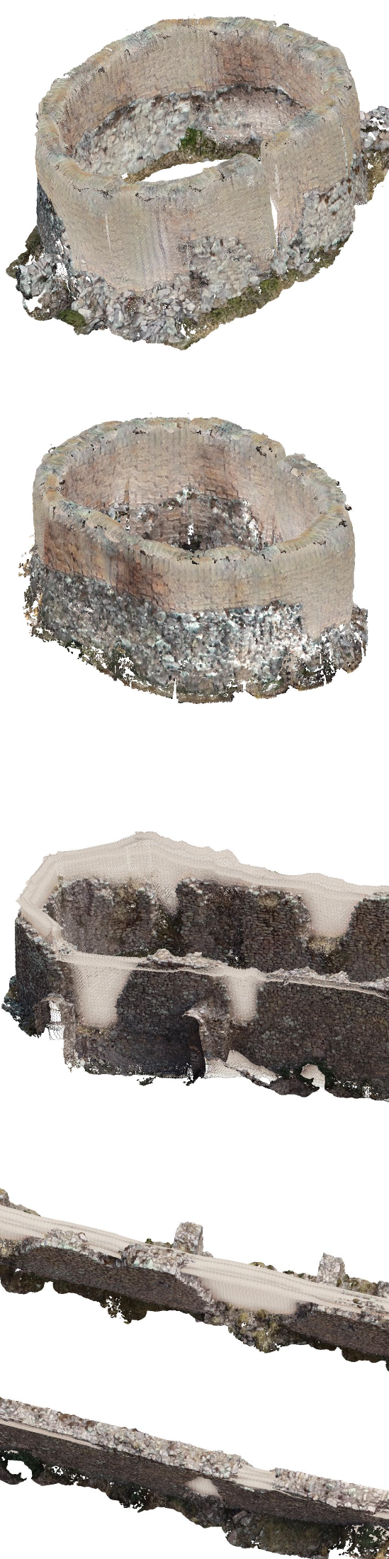}
    \caption{SSII(Patch)}
    \end{subfigure}
    \end{minipage}
    \begin{minipage}{0.13\linewidth}
    \begin{subfigure}{0.9\linewidth}
    \includegraphics[width=\textwidth]{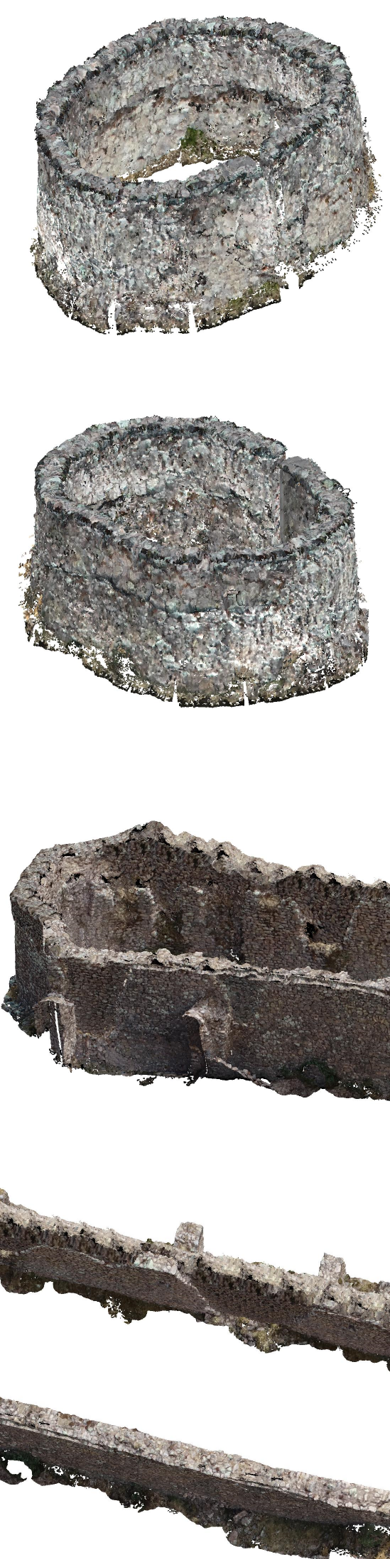}
    \caption{Ours}
    \end{subfigure}
    \end{minipage}
    
  \caption{Point cloud completion results of different methods on \textit{Mawchu Raw}. We overlay completion onto input. More visualization results can be found in the supplemental.}
  \label{fig:pcdVisualization}
\end{figure*}

\subsection{Dataset Construction}

We conduct qualitative and quantitative experiments on hundreds of point cloud datasets of buildings from two real-world archaeological sites, namely Mawchu Llacta and Huamanmarca in Peru which suffer from heavy erosion and are missing a significant portion. Each site is reconstructed from multi-view high resolution UAV images as in \cite{habibTurkey}, consisting of billions of points in total. Point clouds for individual walls/structures are extracted following coarse annotations from archaeologists in our team. To accommodate for efficient learning while not missing out important texture and geometric details, we sample the original point clouds with the proposed improved MCOP at a resolution of 2 centimeters, at which the tallest and biggest structure (as shown in Fig. \ref{titleFig}) are represented within an MCOP image of $384\times 9353$. Since the original point clouds exhibit visible shadows due to lighting or self-occlusion in the input stage, we manually annotate out the shadowed regions in the MCOP images and "deshadow" them by either histogram matching to unshadowed regions or masking out the corresponding regions. Some important statistics for the datasets are listed in supplemental. We name the two raw datasets \textit{Mawchu Raw} and \textit{Huaman Raw}.

\subsection{Quantitative Comparisons}
\begin{figure}[htbp]
  \centering
   \includegraphics[width=1.0\linewidth]{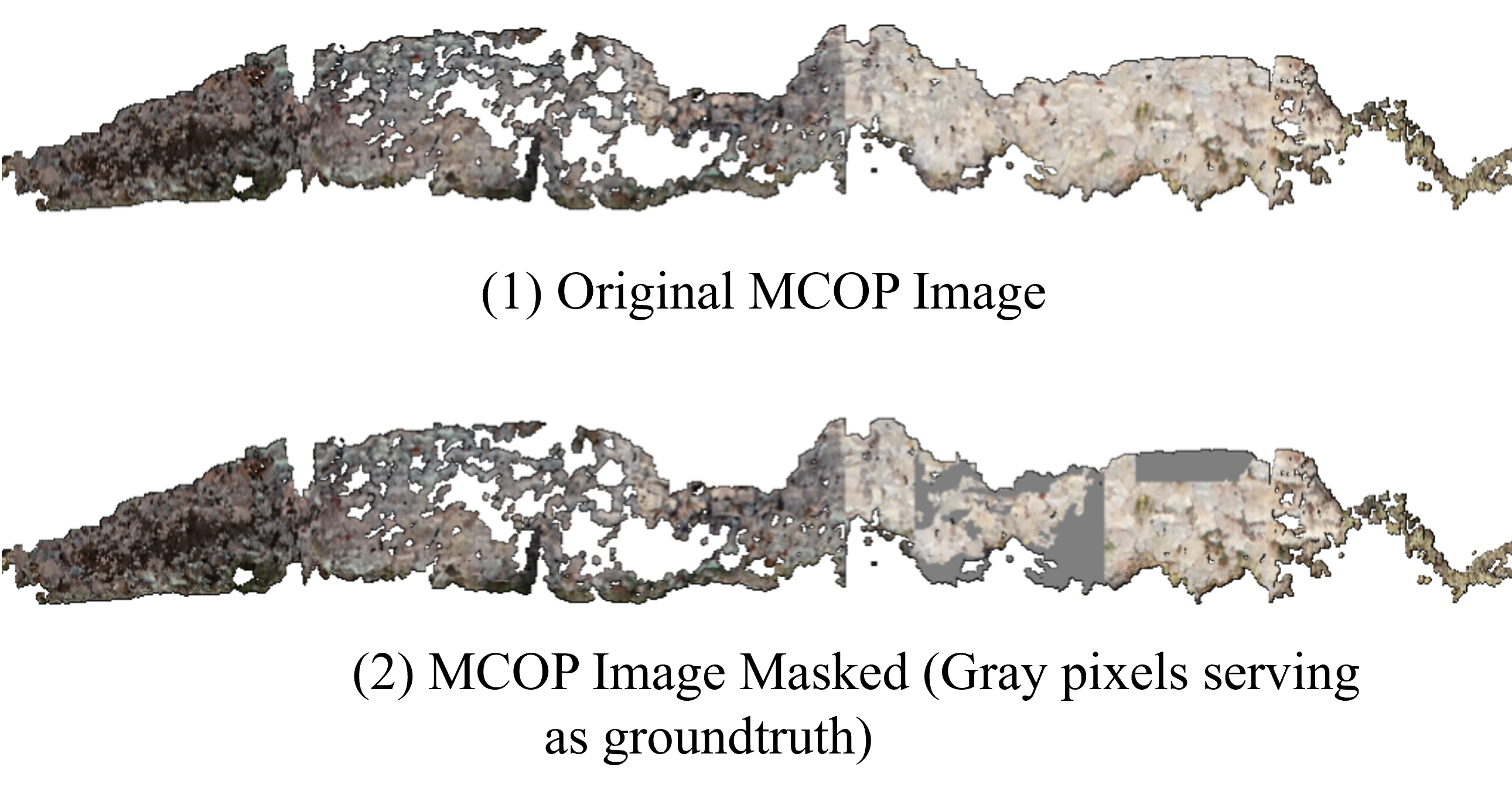}
  
  \caption{Demonstration of the construction of \textit{Mawchu Incomplete} used in evaluation. Using the original MCOP image, we randomly apply small masks from $D_{128}$ onto random locations producing an incomplete MCOP image.}
  \label{fig:erodeIllustration}
\end{figure}

Due to lack of ground truth, we randomly remove sets of points from \textit{Mawchu Raw} to synthesize a more incomplete dataset \textit{Mawchu Incomplete}. The removal is implemented by applying random masks from $D_{128}$ with completeness ratio in between $20\%$ to $80\%$ at random positions in the original MCOP images. As shown in Fig. \ref{fig:erodeIllustration}, the masked out regions could serve as the ground truth for quantitative study when we train on \textit{Mawchu Incomplete} only and evaluate against the masked out regions in \textit{Mawchu Raw}. To fully evaluate the reconstruction quality both in inpainting and point cloud form, we measure the overall RGBD differences of the masked regions, patch/distribution similarity, and Chamfer distance of reconstructed point cloud.

We compare our method on the aforementioned dataset against several state-of-the-art methods. These methods include self-supervised point-cloud approaches (ACL-SPC \cite{ACLSPC} and P2C \cite{p2c}), self-supervised image inpainting methods (UAIR \cite{unsupervisedImageInpainting} and SSII \cite{SSII}) on the MCOP images (i.e., MCOP+UAIR and MCOP+SSII). We also compare against MCOP+SSII(Patch) which uses image-patch self-supervision. ACL-SPC and P2C do not produce colors so we omit those metrics for them.

\begin{table}[h]
	\centering
\begin{tabular}{ccc|c|c}
	\hline
	\multicolumn{3}{c|}{Feature Type} & 
	\multicolumn{1}{c|}{Texture}
	&
	\multicolumn{1}{c}{Geometry}
	\\
	\hline

\multicolumn{1}{c|}{\multirow{6}*{MAE$\downarrow$}} & 
\multicolumn{2}{c|}{ACL-SPC} &
N/A &
0.172
\\
\multicolumn{1}{c|}{~} &
\multicolumn{2}{c|}{P2C} &
N/A &
0.195
\\
\cline{2-5}
\multicolumn{1}{c|}{~} &
\multicolumn{2}{c|}{MCOP+SSII} &
0.206 &
0.143
\\
\multicolumn{1}{c|}{~} &
\multicolumn{2}{c|}{MCOP+UAIR} &
0.259 &
0.149
\\
\multicolumn{1}{c|}{~} &
\multicolumn{2}{c|}{MCOP+SSII(Patch)} &
0.209 &
0.145
\\
\multicolumn{1}{c|}{~} &
\multicolumn{2}{c|}{Ours} &
\textbf{0.189} &
\textbf{0.136}
\\
\hline

\multicolumn{1}{c|}{\multirow{6}*{SSIM$\uparrow$}} & 
\multicolumn{2}{c|}{ACL-SPC} &
N/A &
0.085
\\
\multicolumn{1}{c|}{~} &
\multicolumn{2}{c|}{P2C} &
N/A &
0.048
\\
\cline{2-5}
\multicolumn{1}{c|}{~} &
\multicolumn{2}{c|}{MCOP+SSII} &
0.236 &
0.291
\\
\multicolumn{1}{c|}{~} &
\multicolumn{2}{c|}{MCOP+UAIR} &
0.173 &
0.261
\\
\multicolumn{1}{c|}{~} &
\multicolumn{2}{c|}{MCOP+SSII(Patch)} &
0.213 &
0.278
\\
\multicolumn{1}{c|}{~} &
\multicolumn{2}{c|}{Ours} &
\textbf{0.246} &
\textbf{0.353}
\\
\hline

\multicolumn{1}{c|}{\multirow{6}*{PSNR$\uparrow$}} & 
\multicolumn{2}{c|}{ACL-SPC} &
N/A &
15.06
\\
\multicolumn{1}{c|}{~} &
\multicolumn{2}{c|}{P2C} &
N/A &
13.52
\\
\cline{2-5}
\multicolumn{1}{c|}{~} &
\multicolumn{2}{c|}{MCOP+SSII} &
11.79 &
17.18
\\
\multicolumn{1}{c|}{~} &
\multicolumn{2}{c|}{MCOP+UAIR} &
12.37 &
16.54
\\
\multicolumn{1}{c|}{~} &
\multicolumn{2}{c|}{MCOP+SSII(Patch)} &
11.65 &
17.10
\\
\multicolumn{1}{c|}{~} &
\multicolumn{2}{c|}{Ours} &
\textbf{14.28} &
\textbf{17.87}
\\
\hline

\multicolumn{1}{c|}{\multirow{6}*{FID$\downarrow$}} & 
\multicolumn{2}{c|}{ACL-SPC} &
N/A &
211.69
\\
\multicolumn{1}{c|}{~} &
\multicolumn{2}{c|}{P2C} &
N/A &
331.10
\\
\cline{2-5}
\multicolumn{1}{c|}{~} &
\multicolumn{2}{c|}{MCOP+SSII} &
203.85 &
184.84
\\
\multicolumn{1}{c|}{~} &
\multicolumn{2}{c|}{MCOP+UAIR} &
192.45 &
158.79
\\
\multicolumn{1}{c|}{~} &
\multicolumn{2}{c|}{MCOP+SSII(Patch)} &
179.05 &
169.06
\\
\multicolumn{1}{c|}{~} &
\multicolumn{2}{c|}{Ours} &
\textbf{90.81} &
\textbf{128.41}
\\
\hline

\end{tabular}
	\caption{Quantitative comparison of completion quality using image-level metrics. Our approach has the \textit{best} performance (shown in bold).}
\label{inpaintMetrics}
\end{table}

We first report metrics in image inpainting as shown in Table \ref{inpaintMetrics}. For each completed point cloud, we project it into 2D using our proposed MCOP representation and perform image patch evaluation with the commonly used metrics of MAE, SSIM, PSNR and FID based on $500$ pairs of randomly chosen windows covering the masked out regions. Our method achieves the best performance in both geometry and texture of predicted point cloud in all metrics. 

\begin{table}[h]
	\centering
\begin{tabular}{cc|c|c|c}
	\hline
	\multicolumn{2}{c|}{Method} & 
	\multicolumn{1}{c|}{P$\downarrow$}
	&
	\multicolumn{1}{c|}{C$\downarrow$}
        &
        \multicolumn{1}{c}{CD$\downarrow$}
	\\
	\hline

\multicolumn{2}{c|}{ACL-SPC} &
91.07 &
35.00 &
126.07
\\
\multicolumn{2}{c|}{P2C} &
11.33 &
12.24 &
23.57
\\
\hline

\multicolumn{2}{c|}{MCOP+SSII} &
5.20 &
4.58 &
9.79
\\

\multicolumn{2}{c|}{MCOP+UAIR} &
\textbf{2.90} &
6.29 &
9.20
\\
\multicolumn{2}{c|}{MCOP+SSII(Patch)} &
4.27 &
4.91 &
9.18
\\
\multicolumn{2}{c|}{Ours} &
\underline{3.80} &
\textbf{4.42} &
\textbf{8.22}
\\
\hline

\end{tabular}
	\caption{Quantitative comparison of completion quality using point cloud metrics: Chamfer distance (CD), precision (P), and coverage (C) as used in \cite{ACLSPC} and in centimeters. Our method is best or second best in all metrics.}
\label{pcdMetrics}
\end{table}

We further report the completion quality using the same point cloud metrics as in \cite{ACLSPC}, where we compute the Chamfer distance of prediction $O$ and ground-truth part $GT$ in the masked regions of \textit{Mawchu Incomplete} as follows:
\begin{equation}
    CD(O, GT) = \frac{1}{N_O}\sum_{p\in O}\min_{q\in GT}||p-q||_2 \\
     + \frac{1}{N_g}\sum_{q\in GT}\min_{p\in O} ||q-p||_2
\end{equation}
where $N_o$ and $N_g$ are the number of points in $O$ and $GT$. The first and second term are denoted separatedly as Precision $P$ and $C$ following \cite{ACLSPC}. We achieve best or second best in all metrics.

\begin{figure}[htbp]
  \centering
   \includegraphics[width=\linewidth]{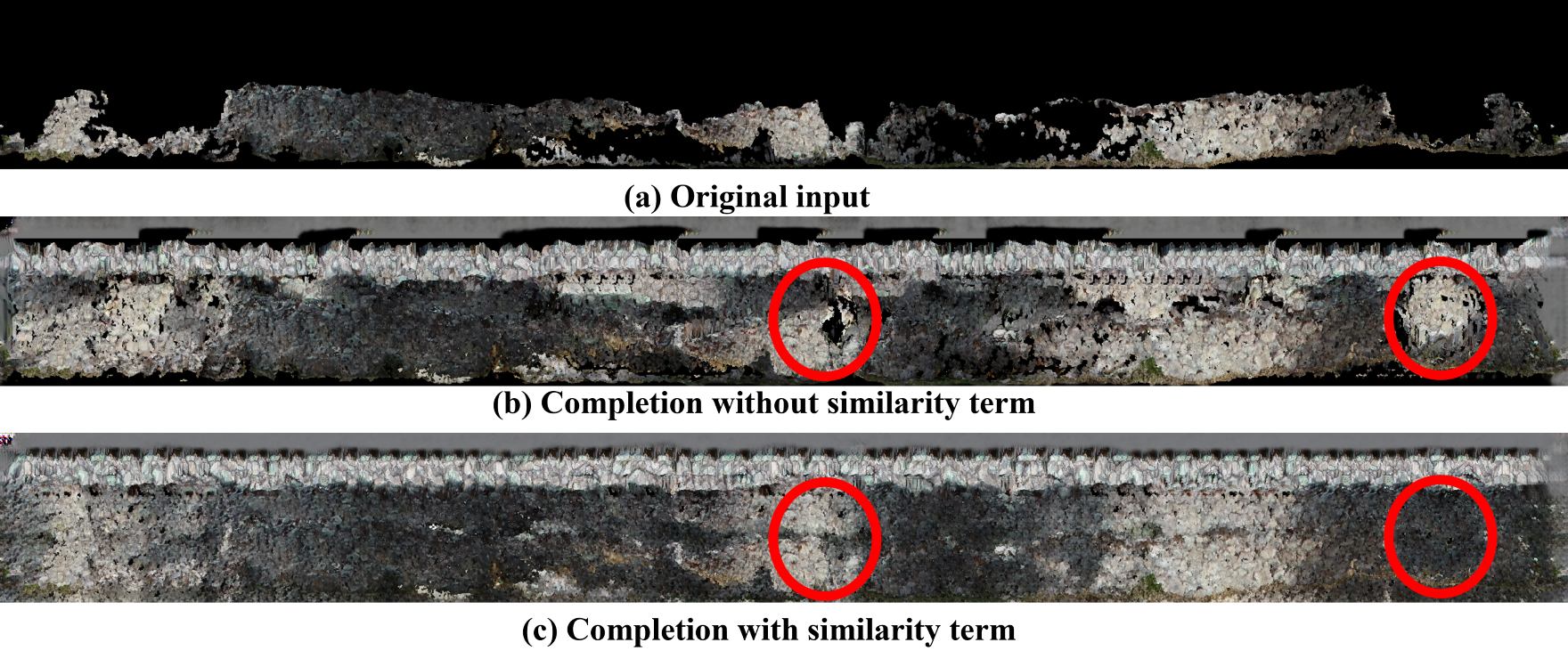}
  
  \caption{Demonstration of need for texture similarity term.}
  \label{fig:textureSimEffect}
\end{figure}

\begin{figure}[htbp]
  \centering

    \begin{subfigure}{1.0\linewidth}
    \includegraphics[width=\textwidth]{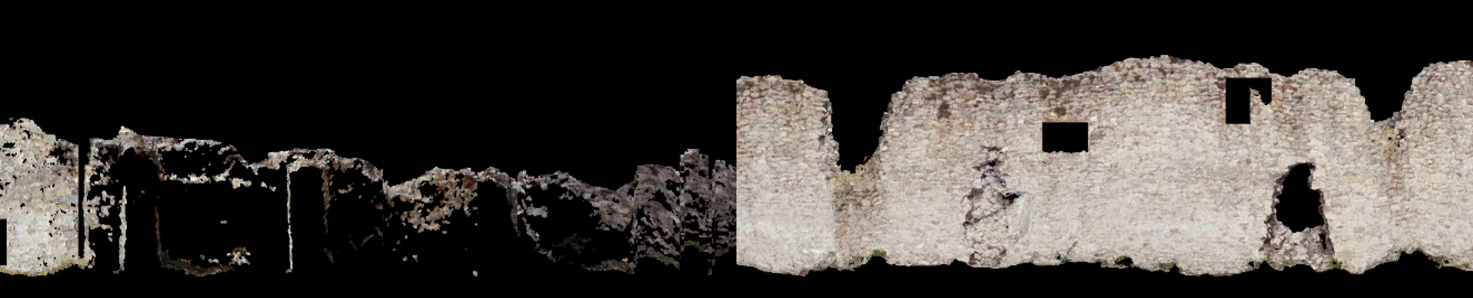}
    \caption{MCOP input}
    \end{subfigure}

    \begin{subfigure}{1.0\linewidth}
    \includegraphics[width=\textwidth]{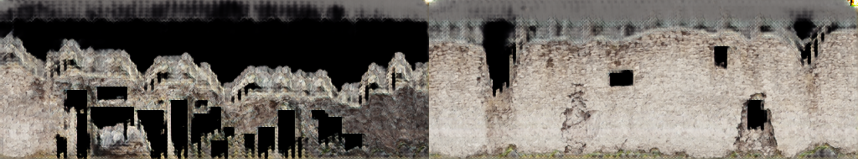}
    \caption{MCOP+UAIR output}
    \label{fig:patch level:uair}
    \end{subfigure}

    \begin{subfigure}{1.0\linewidth}
    \includegraphics[width=\textwidth]{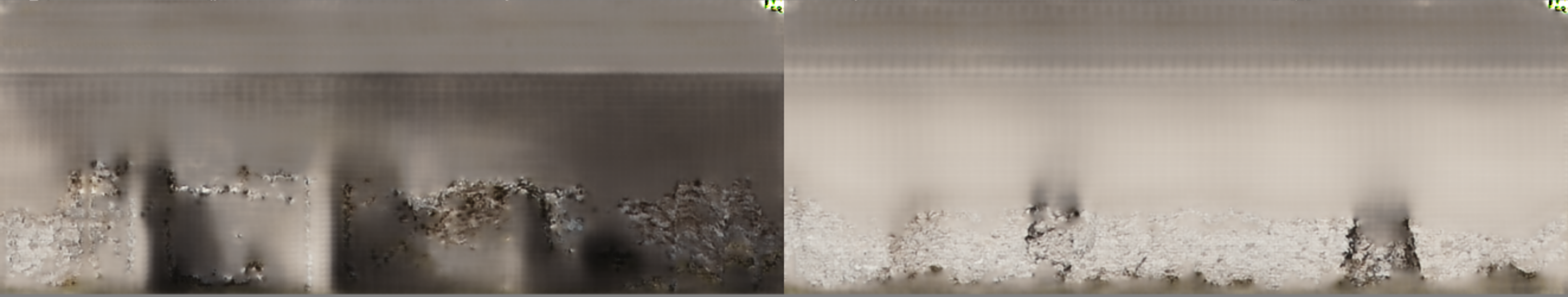}
    \caption{MCOP+SSII output}
    \end{subfigure}

    \begin{subfigure}{1.0\linewidth}
    \includegraphics[width=\textwidth]{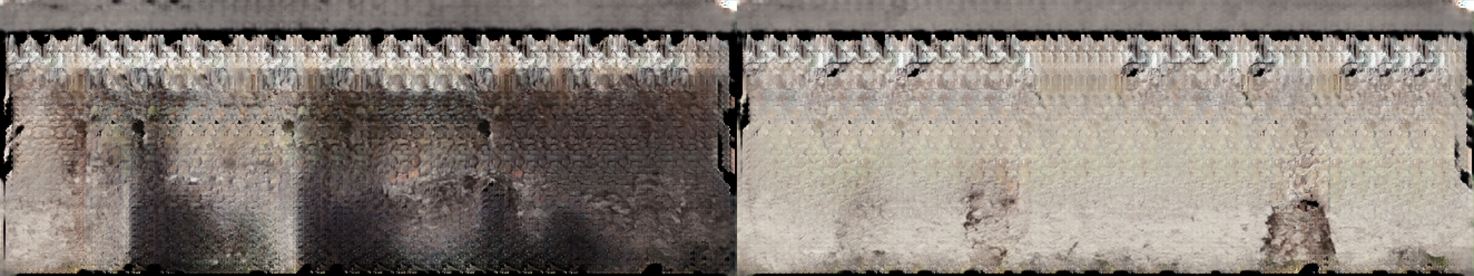}
    \caption{MCOP+SSII(Patch) output}
    \end{subfigure}
    
  \caption{Demonstration of the necessity of applying patch based self supervision for inpainting our MCOP images.
  }
  \label{fig:patch level self supervision}
\end{figure}

\subsection{Ablation Study}

\textbf{MCOP representation.} From Table. \ref{inpaintMetrics} and \ref{pcdMetrics}, methods using MCOP as their base representation generally achieve better performance by a large margin as compared to fully point clouds based ones. The behavior is also interpretable in Fig. \ref{fig:pcdVisualization}, where P2C and ACL-SPC generate noisy and non-uniform points around the input, whereas MCOP efficiently regularizes the output point cloud.
\\
\textbf{Patch-level self supervision.} We demonstrate using patch-level self-supervision in Fig. \ref{fig:patch level self supervision}. Given (a), directly applying UAIR on the completed patches would yield significantly smaller supervision at the higher pixels compared to the rest because of the unbalanced distribution of available pixels (b), for which the completer simply learns to escape adversarial penalty by infilling default pixel values at those positions. SSII also suffers from the unbalanced distribution and tends to produce blurry completions at pixels far from known regions (c). Our patch-based scheme successfully overcomes the unbalanced distribution and helps produce a sharp completion everywhere (d).
\\
\textbf{Texture similarity.} As shown in Fig. \ref{fig:textureSimEffect}, completion without texture similarity constraints produces inconsistent horizontal and vertical patterns at a receptive field larger than the observation window, leading to visual inconsistencies as highlighted in red circles. By incorporating the texture similarity term, the completer is able to maintain consistent completion in both directions at the whole image level.

\section{Conclusions and Future Work}

We propose the first self-supervised point cloud completion pipeline that handles large scale partial point cloud points suffering from missing surfaces, which is common for fragments in archaeological sites. By embedding the original structure in 2D using multiple-center-of-projection (MCOP) images, the approach can handle point clouds of up to millions of points. Patch-based self-supervision with texture similarity terms helps overcome the unbalanced distribution of available pixels resulting from missing surfaces. Results demonstrate our method performs best quantitatively and qualitatively. Nevertheless, our work relies on prior annotation for objects to collect high fidelity MCOP images, which could influence the performance and quality of point cloud completion. As future work, we would like to extend our method to 1) be fully automated to complete input fragments; 2) complete point clouds of general objects at high resolution, and 3) improve shadow elimination method.

{
    \small
    \bibliographystyle{ieeenat_fullname}
    \bibliography{main}
}


\end{document}